\documentclass[11pt]{article}

\usepackage{acl}

\usepackage{times}
\usepackage{latexsym}

\usepackage[T1]{fontenc}

\usepackage[utf8]{inputenc}

\usepackage{microtype}

\usepackage{inconsolata}

\usepackage{graphicx}
\usepackage{xcolor}
\usepackage{adjustbox}
\usepackage{booktabs}
\usepackage{multirow}
\usepackage{subcaption}
\usepackage{makecell}
\usepackage{tabularx}
\usepackage[table]{xcolor}
\usepackage{xurl}
\usepackage{tikz}
\usetikzlibrary{arrows.meta,positioning,shapes.geometric}
\usepackage{pifont}
\usepackage{amsmath}
\usepackage{listings}
\usepackage{xcolor}

\usepackage{mdframed}
\newmdenv[
  backgroundcolor=yellow!7,
  linewidth=0.4pt, % No border
  innertopmargin=3pt, % Space at the start of the box
  innerbottommargin=3pt, % Space at the end of the box
  skipabove=4pt, % Space before the box
  skipbelow=4pt, % Space after the box
  innerleftmargin=3pt,
  innerrightmargin=3pt
]{graybox}
\title{Keyword Matters: Unveiling the Energy Sensitivity of On-Device LLM Prompting}

\author{Ruiyi Tao \\
  Juanita High School \\
  Redmond, Washington, USA \\
  \texttt{taoroy960@gmail.com} \\\And
  Xiaolong Tu \\
  Georgia State University \\
  Atlanta, Georgia, USA \\
  \texttt{xtu1@gsu.edu} \\\And
  Haoxin Wang \\
  Georgia State University \\
  Atlanta, Georgia, USA \\
  \texttt{haoxinwang@gsu.edu} \\}

\begin{document}

\maketitle

\begin{abstract}
Large Language Models (LLMs) are increasingly deployed on mobile and embedded devices to improve privacy and reduce network latency. Yet on-device inference faces a fundamental constraint: high energy consumption on battery-powered, resource-limited hardware. While model compression and runtime acceleration have been widely studied, the effect of \emph{prompt design} on energy efficiency remains underexplored. This paper presents an empirical study of the relationship between prompt wording and energy consumption for on-device LLMs. Using real power measurements collected on a smartphone, we quantify how linguistic features, particularly imperative keywords and instruction structure, affect decoding length and total energy. Our results show consistent energy differences across verbs and tasks, indicating that prompt engineering is a lightweight lever for improving energy efficiency. 
% Code and tutorials are available at \url{https://github.com/RuiyiTao/KeywordMatters}.
\end{abstract}

\begin{figure*}[t]
    \centering
    \includegraphics[width=\linewidth]{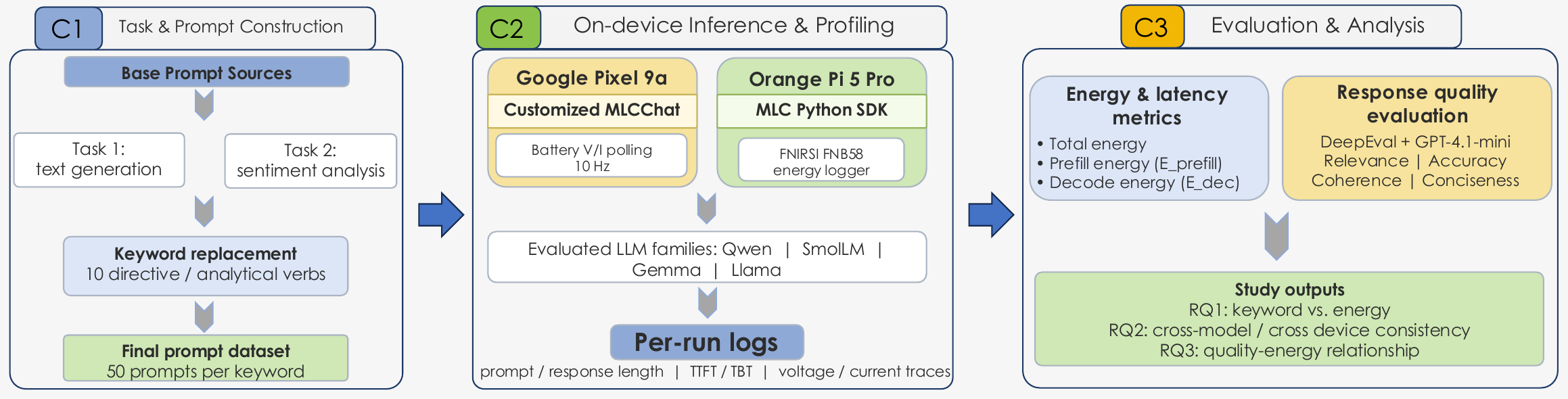}
    \caption{Overview of the experimental pipeline. Prompts are constructed from text generation and sentiment analysis tasks through keyword replacement, then executed on two edge platforms, Google Pixel 9a | Android 15 and Orange Pi 5 Pro | Ubuntu 22.04, for on-device inference and energy profiling.}
    \label{fig:pipeline}
\end{figure*}

\section{Introduction}

Large Language Models (LLMs), such as GPT-4, LLaMA 3, and DeepSeek, have demonstrated remarkable capabilities in text generation, reasoning, and problem solving. They now power applications ranging from virtual assistants and coding copilots to educational and creative tools. To support these workloads, most commercial LLMs are currently deployed in cloud environments, where large-scale computing clusters provide sufficient memory, processing, and cooling resources to handle billions of parameters and real-time user requests \cite{llama3_herd,phi3tech}.

However, cloud-based inference introduces growing concerns about data privacy, latency, and energy cost. Sensitive user data must travel across networks to remote servers, creating potential privacy risks and increasing response delays. As a result, researchers and industry developers have begun exploring on-device LLM deployment, running compact models directly on mobile phones, embedded systems, and IoT devices \cite{mlcllm2023,tensorrtllm,han2016deepcompression,hinton2015distillation}. This emerging trend offers improved privacy and offline availability, but also exposes a fundamental constraint: limited energy and computing resources.

Unlike data centers with continuous power supply and active cooling, edge devices are battery-powered and thermally constrained. The high computational demand of autoregressive token generation leads to substantial energy draw on CPUs, GPUs, or NPUs, shortening device battery life and limiting sustained performance \cite{schwartz2019greenai,strubell2019energy}. As model inference scales with prompt length, decoding steps, and reasoning depth, even small inefficiencies in model execution or input design can result in significant energy overhead. Therefore, understanding and reducing energy consumption has become a crucial step toward enabling practical and sustainable on-device intelligence.

While prior research has optimized model architectures through quantization, pruning, and distillation, the effect of prompt design on energy consumption remains underexplored. Recent works suggest that different prompt wordings can trigger distinct reasoning paths, token lengths, and attention activations, potentially altering power usages. This raises an intriguing question: \textit{Can prompt engineering serve as a lightweight mechanism for energy-aware control during on-device LLM inference?}

In this work, (i) We present a controlled measurement study of keyword-level prompt effects on decoding energy. (ii) We define a measurement protocol and metrics (TTFT, TBT, energy split into prefill/decoding) and report cross-model, cross-task trends. We analyze how different linguistic patterns, especially keywords, affect runtime behavior and power draw. Our findings highlight that prompt formulation is not only a semantic or stylistic factor, but also an energy-relevant one.

\section{Related Work}

\subsection{LLM Optimization and On-Device Deployment}
Recent advances in system and algorithm co-design have significantly reduced LLM inference cost. FlashAttention reorders attention computation to minimize memory access and achieve GPU speedups~\cite{dao2022flashattention}, while vLLM introduces PagedAttention to mitigate KV-cache fragmentation and improve long-context throughput~\cite{kwon2023vllm}. Speculative decoding accelerates generation through draft-and-verify prediction~\cite{leviathan2023speculative}.
Model-side compression complements these efforts. For example, QLoRA enables memory-efficient fine-tuning on low-bit models~\cite{dettmers2023qlora}.

Beyond cloud servers, compact transformer architectures such as MobileLLM~\cite{liu2024mobilellm} demonstrate the feasibility of sub-billion-parameter LLMs on edge devices. Frameworks like MLC-LLM and TensorRT-LLM further optimize execution pipelines for mobile and embedded NPUs. Meanwhile, sustainability studies such as CarbonTracker~\cite{anthony2020carbontracker} and the BLOOM carbon report~\cite{luccioni2023bloom} provide tools for quantifying energy and emissions in AI training and inference. However, most prior work focuses on hardware or model efficiency rather than user-driven linguistic effects on energy consumption.

\subsection{Prompt Engineering and Energy Awareness}
Prompting strategies directly influence reasoning depth and token generation. Chain-of-Thought~\cite{wei2022cot} enhances multi-step reasoning at the cost of longer decoding sequences. Automatic Prompt Engineering (APE)~\cite{zhou2022ape} and Active-Prompt~\cite{diao2023activeprompt} search or select effective prompts to guide model behavior. While these studies focus on improving accuracy, efficiency, or controllability, few consider their impact on energy usage.
Our work extends this line by empirically measuring how imperative keyword choices in prompts affect \textit{prefill} and \textit{decode} energy consumption on a smartphone, introducing a sustainability perspective to prompt engineering and bridging the gap between linguistic design and device-level energy efficiency.

\section{Background and Motivation}
\label{section:3}

\subsection{Background}
Large Language Models (LLMs) such as ChatGPT, Gemini, and Claude rely on massive cloud-based GPU and TPU clusters for real-time inference. While this setup enables scalability and continuous updates, it raises concerns about \textbf{privacy}, \textbf{latency}, and \textbf{energy consumption}. Global data centers hosting AI workloads are projected to consume nearly 1{,}000~TWh of electricity annually by 2030.

To mitigate these challenges, recent progress in quantization, pruning, and efficient transformer architectures has enabled \textit{on-device LLMs}, compact models such as \texttt{Phi-3-mini}, \texttt{Qwen-1.5}, and \texttt{LLaMA~3.2~1B/3B}---that can run locally on smartphones or embedded NPUs. These on-device deployments improve privacy and responsiveness but introduce a new bottleneck: \textbf{energy efficiency}. Autoregressive inference involves two main phases: \textit{prefill} and \textit{decode}, where the latter dominates both runtime and power usage. Empirical profiling shows that total energy scales almost linearly with output length and decoding latency, making energy consumption a critical constraint for edge deployment.

\subsection{Motivation}
Prior efforts to improve energy efficiency have focused on hardware- or model-level techniques, such as quantization, pruning, or kernel optimization, which require retraining and platform-specific tuning. However, a lightweight and overlooked alternative lies in \textbf{prompt design}. Linguistic choices such as ``\textit{generate},'' ``\textit{explain},'' or ``\textit{list}'' can alter reasoning depth, token length, and activation patterns---thereby affecting energy use even under the same model and hardware configuration.

Despite growing evidence that prompt phrasing influences computation, existing studies are largely theoretical or simulation-based, lacking validation on real devices. This work fills that gap by providing the first empirical analysis linking \textit{imperative keyword choices} to measured power consumption during on-device LLM inference. Our findings highlight prompt engineering as a simple yet effective lever for achieving \textit{energy-efficient and privacy-preserving} intelligence at the edge.

\section{Experimental Design}
\label{section:4}

\subsection{Research Objective and Questions}

Building on Section~\ref{section:3}, our objective is to quantify how linguistic variations, particularly the choice of imperative \textit{keywords}, affect the energy consumption of on-device LLMs.
% We hypothesize that creative or analytical verbs (e.g., ``create,'' ``explain,'' ``generate'') induce longer, more complex decoding and thus higher energy usage than directive verbs (e.g., ``list,'' ``rewrite,'' ``identify'').
This study is guided by three research questions:
\begin{itemize}
    \item \textbf{RQ1:} How do different prompt keywords affect the overall energy consumption of on-device LLM inference?
    \item \textbf{RQ2:} Are the observed keyword-driven energy patterns consistent across different LLM models and devices?
    \item \textbf{RQ3:} How is response quality affected by prompt keywords, and how does it correlate with energy consumption?
\end{itemize}

\subsection{Hardware Setup, Model and Experimental Pipeline}

\textbf{Hardware Setup:} Experiments were conducted on an Orange Pi 5 Pro running Ubuntu 22.04 and a rooted Google Pixel~9a. 
These two devices each have different capabilities that became apparent in the study. The Orange Pi excelled in prefill, completing its prefill phase in an average of 115ms, compared to the Pixel 9a, which completed its prefill phase in an average of 87,000ms. However, the decode rate was similar across devices, 114ms TBT for Orange Pi, and 100ms TBT for Pixel 9a. 
To reduce interference from unrelated system activity, all measurements on the Orange Pi were performed in multi-user mode. On the Pixel~9a, CPU frequencies were fixed at 1328~kHz for Cores~0--3, 1418~kHz for Cores~4--6, and 2294~kHz for Core~7. To maintain stable power measurements, thermal throttling, adaptive brightness, and background services were disabled, and screen brightness was fixed at the minimum level to reduce display-related noise.

\textbf{Models and Experimental Pipeline:} The end-to-end measurement workflow is illustrated in Figure~\ref{fig:pipeline}. We evaluated models from the Qwen, SmolLM, Gemma, and Llama families to ensure that our observations are consistent across architectures. Most selected models were matched to have similar parameter sizes and quantization settings, while one smaller model was additionally included to study the effect of model size. All models except the Gemma model are instruct models. The Gemma model was selected to test how a different type of model would be affected by different keywords. 

\subsection{Tasks, Prompts, and Data Collection}

To relate linguistic structure to energy consumption, we evaluate two representative NLP tasks capturing open-ended and analytical behavior:

\textbf{Text Generation:} Prompts adapted from the Alpaca-GPT4 dataset~\cite{alpaca_gpt4} (creative writing, explanation). 50 prompts were chosen, and the keyword was replaced with 10 selected directive verbs. Previously, we chose 50 different prompts for each keyword instead of using the same 50 prompts for each keyword. However, due to the inherent biases in the usage of each keyword in the Alpaca-GPT4 dataset, this led to results that were suspiciously consistent across environments. The text generation prompts used in this paper only differ from each other by keyword, isolating its impact, also preserving the inherent randomness of language models. 

\textbf{Sentiment Analysis:} Prompts synthesized using Yelp review data~\cite{yelp_open_dataset} with analytical verbs (\textit{determine}, \textit{label}, etc.), representing short, deterministic reasoning. These prompts use the same 50 yelp reviews, combined with a short sentence that prompts binary sentiment analysis. Similarly to text generation, this was also done to isolate the impact of keywords. 

\begin{figure}[t]
    \centering
    \includegraphics[width=\linewidth]{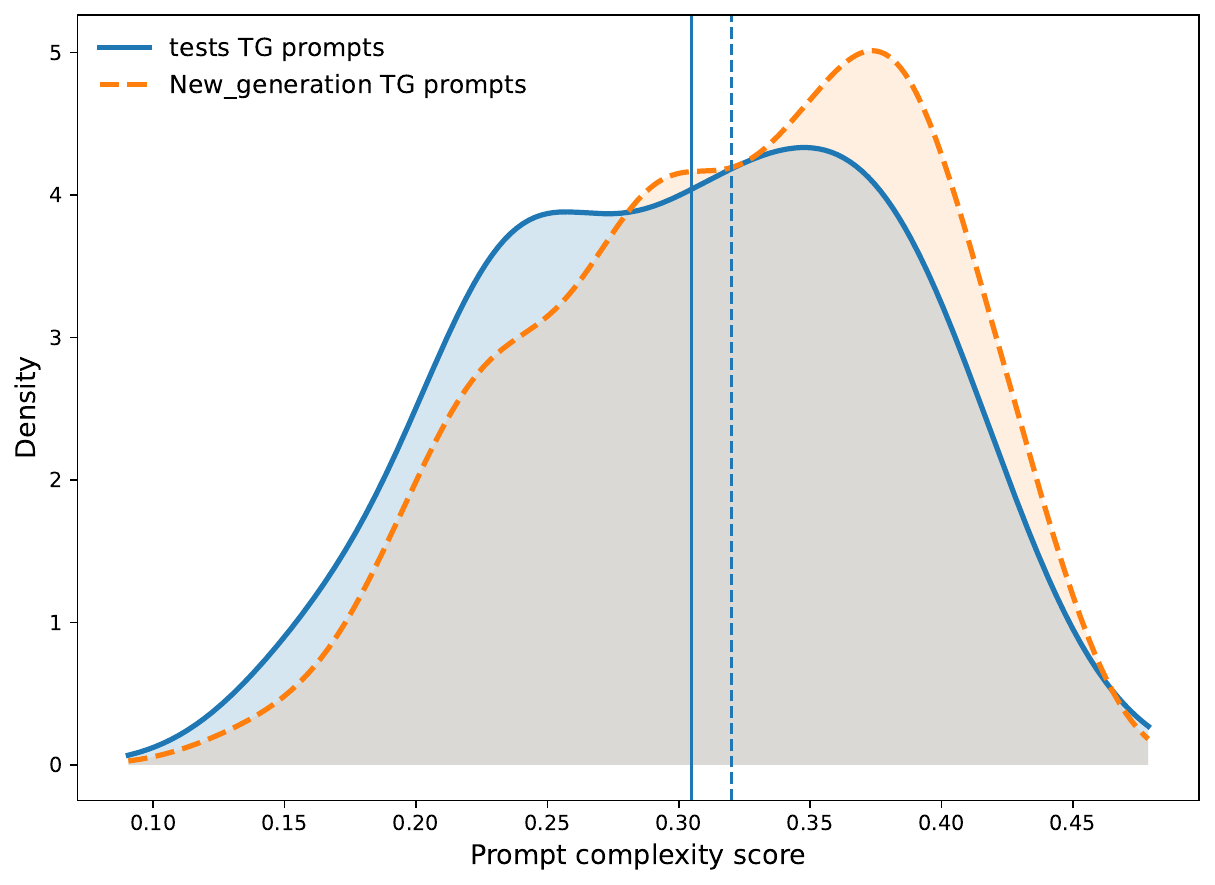}
    \caption{Prompt-complexity distribution comparison between the original and keyword-replaced text-generation prompt dataset.}
    \label{fig:prompt_complexity_dist}
\end{figure}

\textbf{Prompts:} For each keyword, we selected 50 prompts, resulting in 1{,}000 prompts in total (500 per task) and 5{,}000 prompt-model runs across 5 models. To verify that keyword replacement does not substantially change prompt difficulty, we evaluated both the original and rewritten Text Generation prompts using the NVIDIA prompt-task-and-complexity classifier \cite{nvidia_prompt_complexity_classifier}. As shown in Figure \ref{fig:prompt_complexity_dist}, the two score distributions largely overlap, suggesting that the rewritten prompts preserve the overall complexity profile of the original ones. Quantitatively, the matched original prompts have a mean complexity score of 0.3048, compared to 0.3199 for the rewritten prompts, corresponding to a small average shift of +0.0151 on a 0--1 scale. Furthermore, after averaging across rewritten variants, the prompt-level scores remain highly correlated with the original prompts (Pearson $r = 0.942$, Spearman $\rho = 0.944$). Together, these results show that keyword replacement introduces only a minor change in classifier-assigned complexity while preserving the relative difficulty structure of the prompt set.

\textbf{Data Collection:} For the Orange Pi 5 Pro, we used the Python API to run the models and the FNIRSI FNB58 to collect energy data. The output was parsed using a third-party data logger~\cite{fnirsi_logger}.
For the Pixel 9a, we ran a customized \texttt{MLCChat} that integrates logging into the inference pipeline by measuring the power draw from the battery while the phone is discharging. 
Responses are capped at 1{,}000 tokens to stop infinite loops occasionally observed in small quantized models.

Per-run metrics include: \textit{prompt length} ($L_p$), \textit{response length} ($L_r$), \textit{TTFT}, \textit{TBT}, and instantaneous \textit{voltage/current} traces sampled at 10~Hz.
Total energy is computed as:
\[
E = \sum_{i=1}^{N} I_i V_i \Delta t, \qquad \Delta t = 0.1~\mathrm{s},
\]
and logged separately for \textit{prefill} ($E_\text{prefill}$) and \textit{decoding} ($E_\text{dec}$).
To prevent memory accumulation, the model instance is reloaded before each run.

\subsection{Accuracy Evaluation}

To evaluate response quality, we used the DeepEval Python library~\cite{deepeval_repo} with GPT-4.1-mini as an LLM judge to perform G-Eval on each input--output pair~\cite{deepeval_geval_docs}. Model responses were assessed according to the following criteria:
\begin{enumerate}
\item\textbf{Relevance:} whether the output meaningfully addresses the questions, tasks, or instructions expressed in the input prompt.

\item\textbf{Accuracy:} whether the factual claims in the output are correct, verifiable, and consistent with established knowledge or information explicitly provided in the input.

\item\textbf{Coherence:} whether the output is logically organized, internally consistent, and easy to follow as a unified response.

\item\textbf{Conciseness:} whether the output conveys the required information efficiently, without unnecessary detail beyond what is needed to satisfy the intent of the input.
\end{enumerate}

Although the judge model may implicitly consider factual accuracy across multiple criteria, we found that this overlap has minimal effect on the significance of the overall score.

\begin{figure*}[t]
    \centering
    \includegraphics[width=\textwidth]{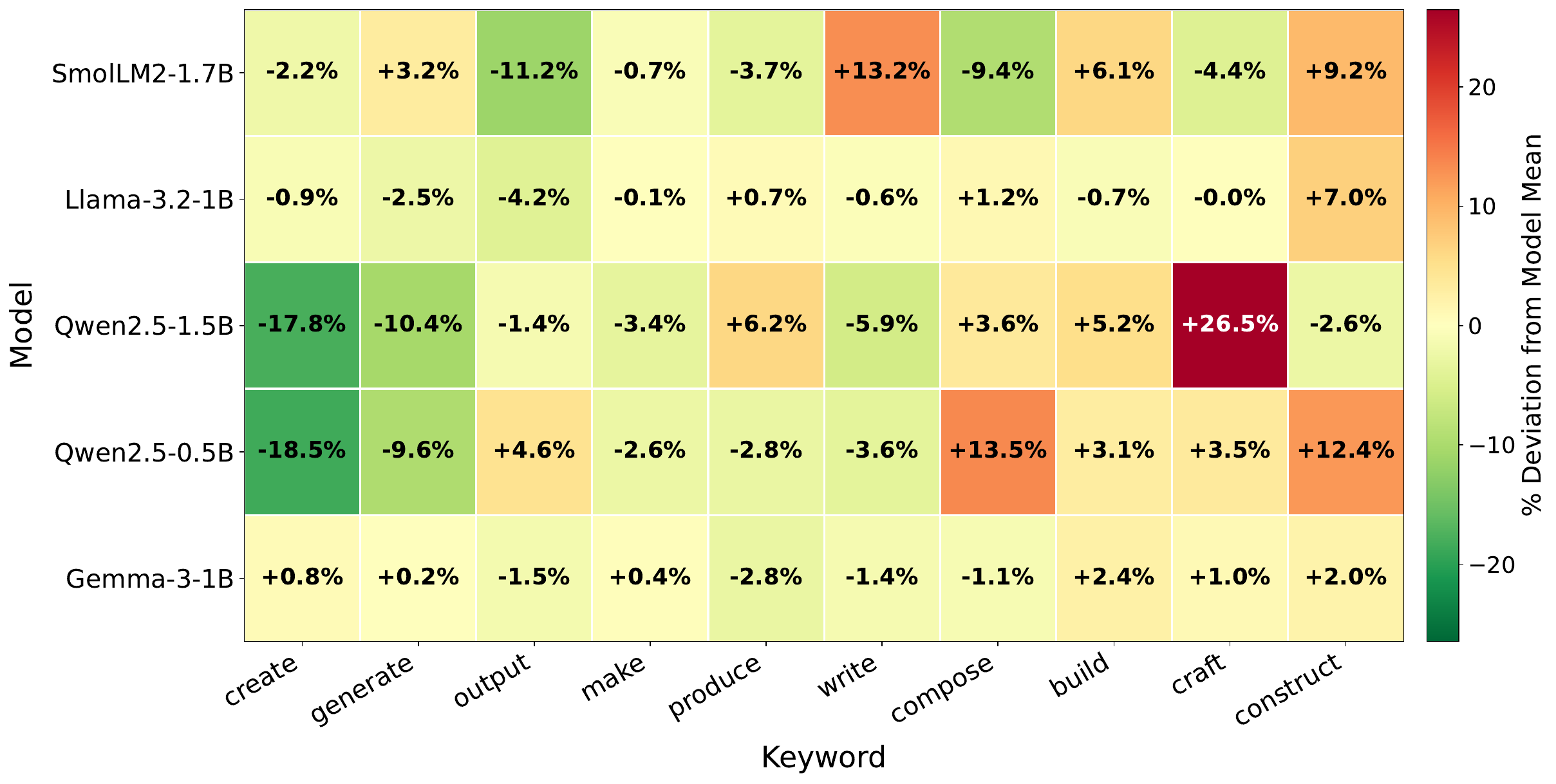}
    \caption{\textbf{Text generation} decode energy deviation from model mean (\%) across classification-related keywords on the Orange Pi 5 Pro. Values represent per-keyword deviation from each model's mean decode energy.}
    \label{fig:tg_mean_energy}
\end{figure*}

\begin{figure*}[t]
    \centering
    \includegraphics[width=\textwidth]{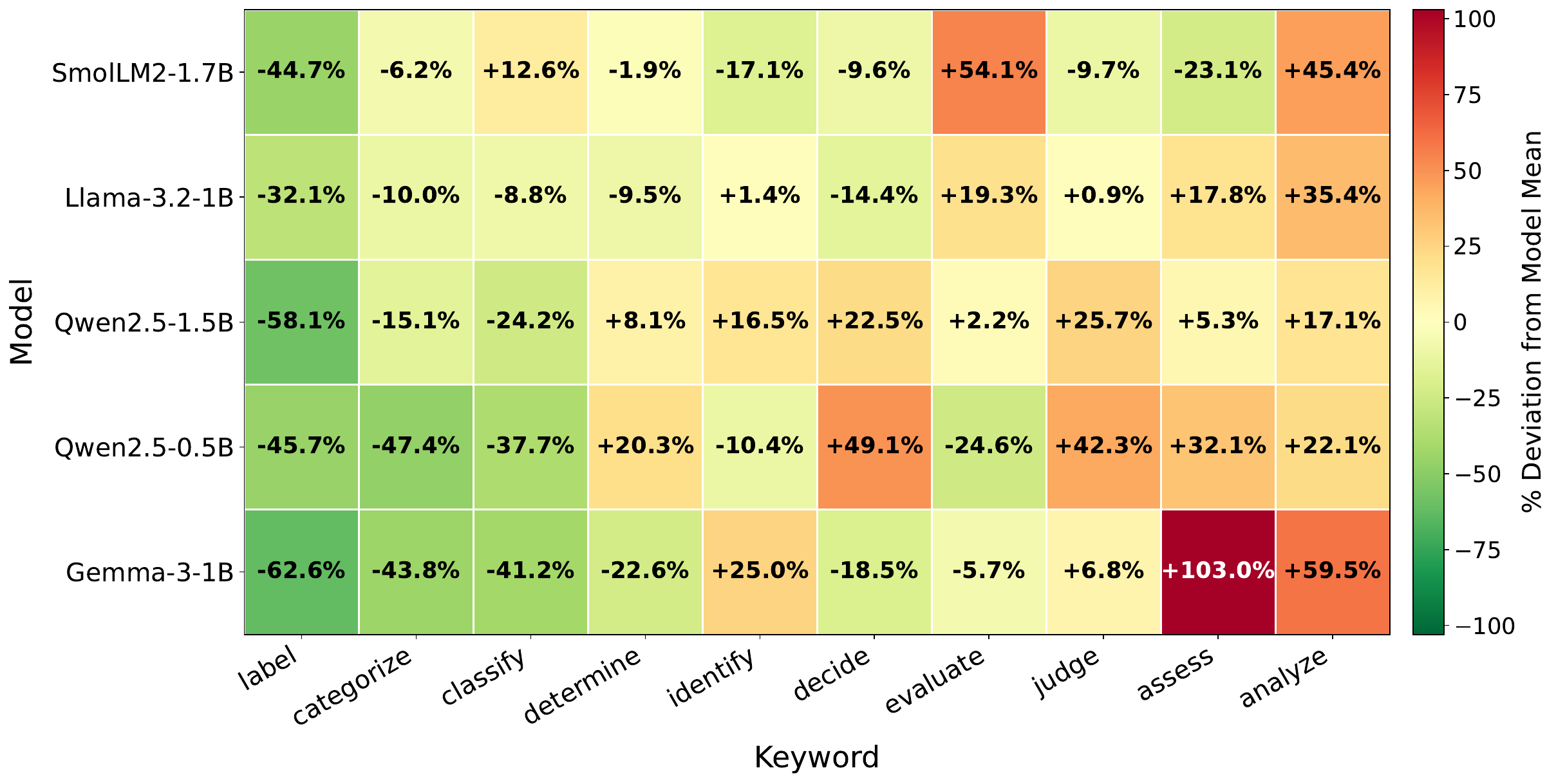}
    \caption{\textbf{Sentiment analysis} decode energy deviation from model mean (\%) across classification-related keywords on the Orange Pi 5 Pro. Values represent per-keyword deviation from each model's mean decode energy.}
    \label{fig:sa_mean_energy}
\end{figure*}

\begin{figure}[t]
    \centering
    \includegraphics[width=\linewidth]{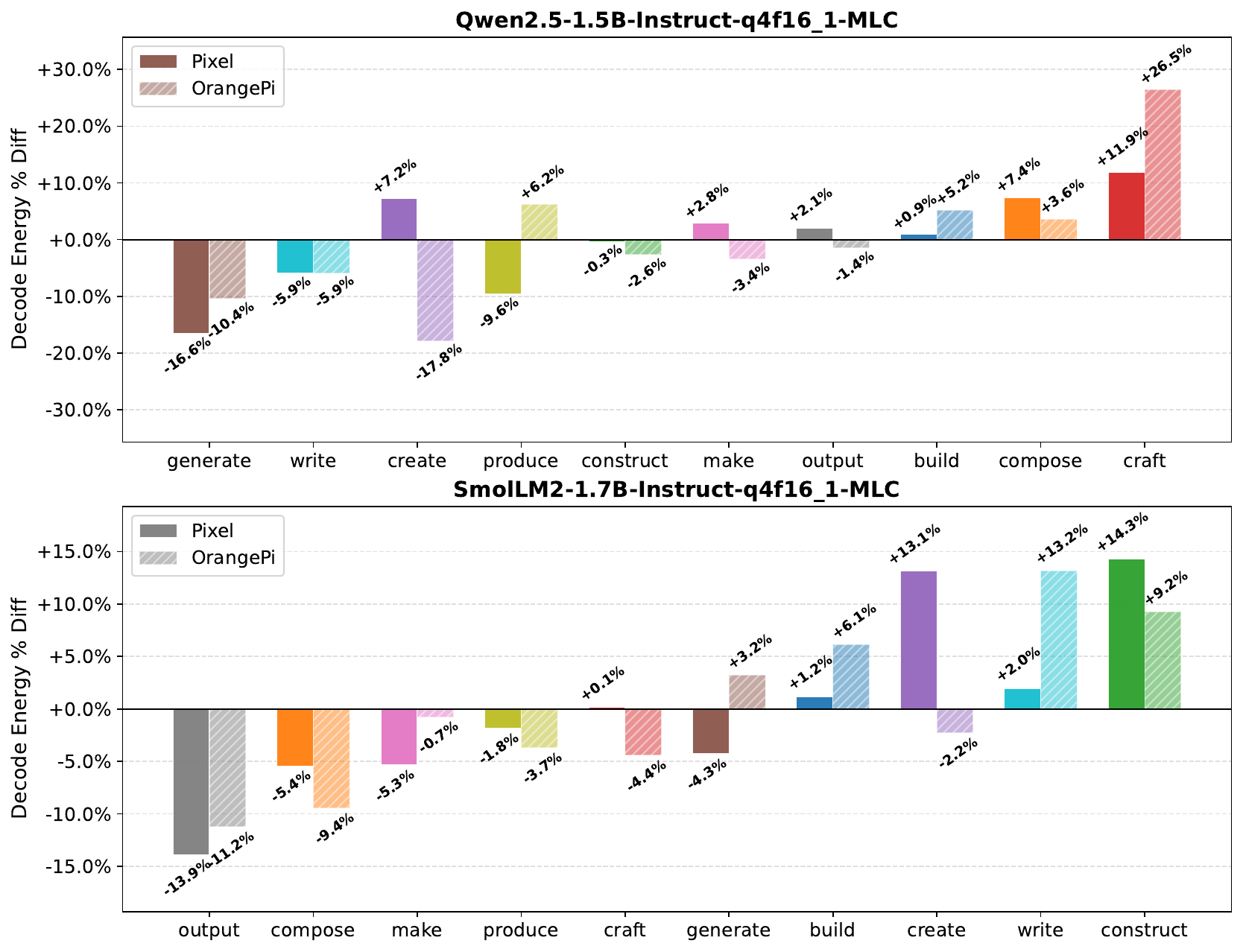}
    \caption{\textbf{Text generation} decode energy deviation from model mean for Pixel 9a and Orange Pi 5 Pro, shown for Qwen2.5-1.5B and SmolLM2-1.7B.}
    \label{fig:pixel_tg_compare}
\end{figure}
\begin{figure}[t]
    \centering
    \includegraphics[width=\linewidth]{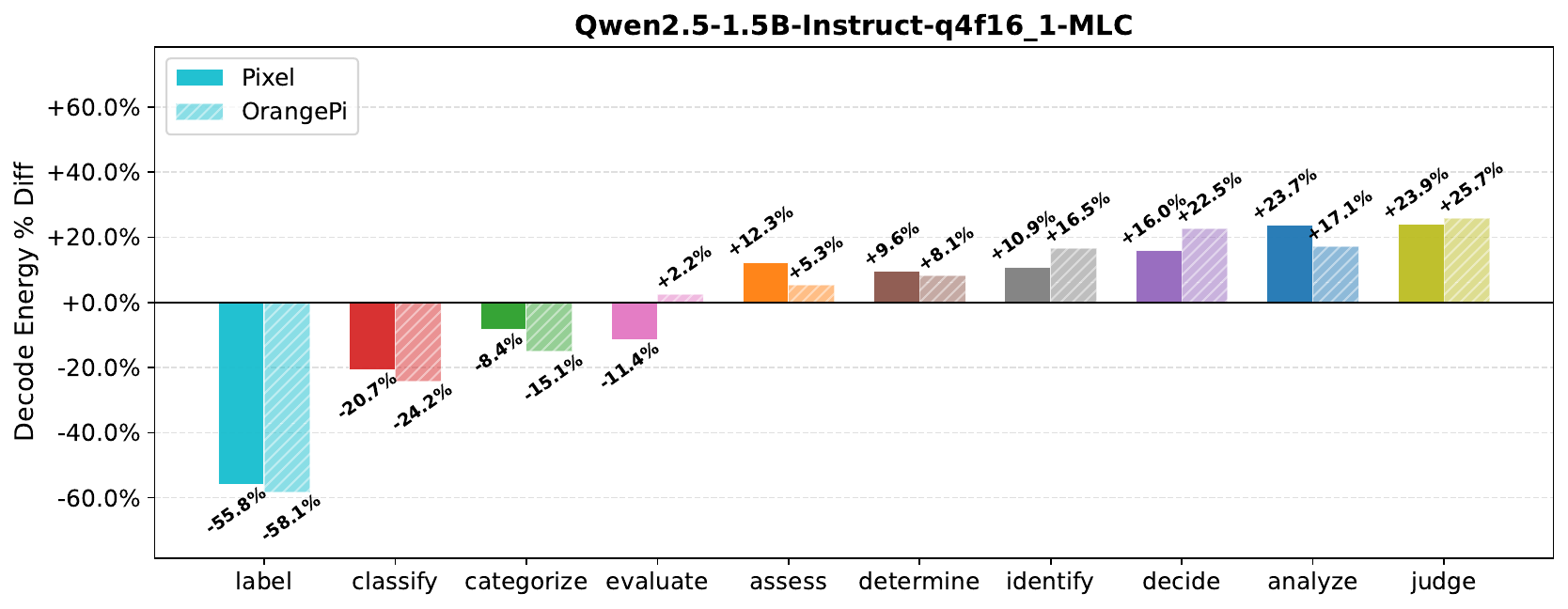}
    \caption{\textbf{Sentiment analysis} decode energy deviation from model mean for Pixel 9a and Orange Pi 5 Pro, shown for Qwen2.5-1.5B.}
    \label{fig:pixel_sa_compare}
\end{figure}

\begin{figure}[!t]
    \centering
    \includegraphics[width=\linewidth]{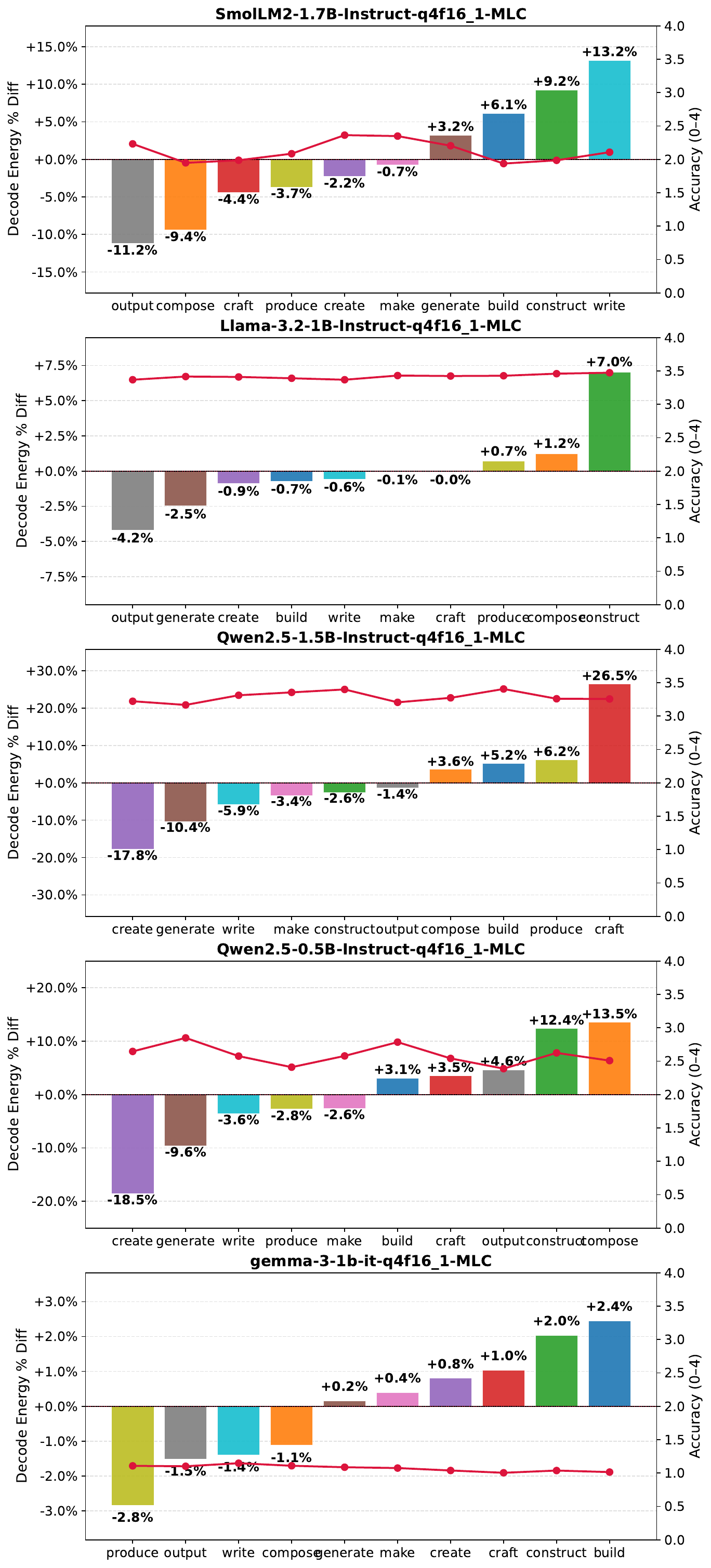}
    \caption{\textbf{Text generation} decode energy \% deviation from model mean and output accuracy on the Orange Pi 5 Pro.}
    \label{fig:tg_accuracy}
\end{figure}
\begin{figure}[!t]
    \centering
    \includegraphics[width=\linewidth]{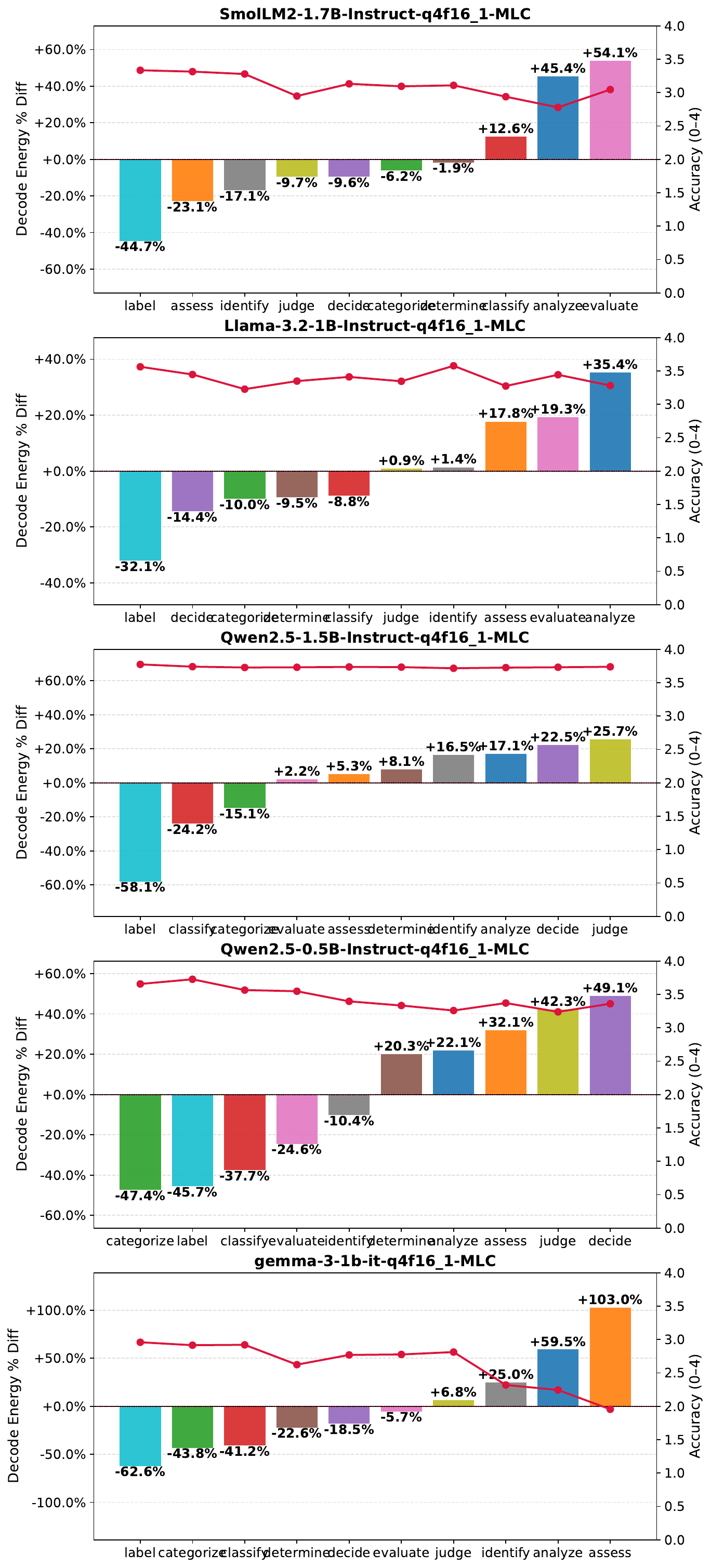}
    \caption{\textbf{Sentiment analysis} decode energy \% deviation from model mean and output accuracy on the Orange Pi 5 Pro.}
    \label{fig:sa_accuracy}
\end{figure}

\section{Empirical Study and Results}

Building upon the experimental framework described in Section~\ref{section:4}, this section presents empirical findings addressing our three research questions (RQ1--RQ3).
We analyze how linguistic variations, particularly keywords, influence the energy consumption of on-device LLMs across 5 models and two task types (text generation and sentiment analysis).
Unless otherwise specified, all energy values refer to the decoding phase ($E_{\text{dec}}$).

All figures show how the mean energy consumption for each keyword compares to the mean energy consumption across the whole model, represented as a percentage deviation.

\subsection{RQ1: How do different prompt keywords affect the overall energy consumption of on-device LLM inference?}
\subsubsection{Text Generation}

Figure~\ref{fig:tg_mean_energy} reveals that using an energy efficient keyword can save up to 18.5\% energy during the decode phase of text generation tasks (``create'' keyword in Qwen 0.5B). Meanwhile, an inefficient choice of keyword can increase consumption by up to 26.5\% (``craft'' keyword in Qwen 1.5B).
In both Qwen models, the top performing keyword, ``create,'' saved around 18\% energy in the decode phase. Compare this to Llama, which saved only 4.2\% energy with the ``output'' keyword. 

\begin{graybox}
\underline{Insight 1:} \textit{Certain model families, like Qwen, are more sensitive to changes in keywords, leading to larger energy savings}.
\end{graybox}

Of these models, the order from most to least sensitive is: Qwen, SmolLM, Llama, Gemma. 
Gemma had the lowest keyword sensitivity, but this is because Gemma's responses were low quality (see Figure~\ref{fig:tg_accuracy}) and often devolved into endless repetition, quickly reaching the token limit. Since the lengths of many responses were the same, at token limit, there is a relatively small deviation in energy consumption.
In the second least sensitive model, Llama, aside from the keyword ``construct'', which led to a 7\% increase in energy consumption, the rest of the keywords consumed similar amounts of energy, deviating from each other by less than 6\%.

\subsubsection{Sentiment Analysis}

Figure~\ref{fig:sa_mean_energy} shows that there is more promise to reduce energy consumption in sentiment analysis. Across all 5 models, the keyword ``label'' performed very well, saving between 32.1\% and 62.6\% decode energy from the mean in all models, suggesting that it could be heavily optimized in other language models as well.

Energy savings in sentiment analysis are much greater than in text generation when viewing percentages. This is because sentiment analysis generates shorter responses, so small changes in response length can lead to large percentage changes.
Aside from the ``label'' keyword, the ``categorize'' keyword also performed better than average in all 5 models, but it especially shined in Qwen 0.5B.

\subsection{RQ2: Are the observed keyword-driven energy patterns consistent across different LLM models and devices?}
\subsubsection{Consistency across models}

Using the pairwise Spearman coefficient of the ranks of keywords across models, we get an average coefficient of 0.153 in text generation, and an average $\rho = 0.388$ in sentiment analysis, showing that there is very little correlation across models. However, some patterns stand out. 

The two Qwen models had a Spearman correlation of $\rho = 0.588$ in text generation, and $\rho = 0.867$ in sentiment analysis. In addition, the top three keywords for energy efficiency were ``create,'' ``generate,'' and ``write,'' in this order for both models. 
In text generation, for all keywords but 3 (``craft,'' ``construct,'' and ``compose''), they deviated between models by an average of only 3.4\%. 

\begin{graybox}
\underline{Insight 2:} \textit{In the same model family, energy consumption is consistent with the keyword used}.
\end{graybox}

Despite the insignificant correlation, a trend in all 5 models in text generation is that keywords like ``generate'' and ``output'' that are robotic in nature seem to perform well in all models (average deviation by -4.6\% and -2.7\% respectively, and ranks 2 and 3 overall across all models), suggesting that the models may be optimized for robotic keywords. However, the ``create'' keyword outperformed both with an average deviation by -7.7\%.
The worst performing keyword was ``construct,'' with an average deviation of +5.6\% across all models. 

In sentiment analysis, the best performing keyword was ``label,'' which performed well in all 5 models, with an average percentage deviation of -48.64\%. Other keywords that performed well include ``categorize'' (-24.5\%), and ``classify'' (-19.86\%).
Compared to that, the keywords that performed the worst were ``analyze'' (+35.9\%) and ``assess'' (+27.02\%). 
The keywords that performed well all instructed the model to identify rather than analyze. 

\begin{graybox}
\underline{Insight 3:} \textit{Keyword sentiment has a tangible impact on energy consumption in sentiment analysis tasks; keywords that prompt deep thinking in humans also prompt more complex thinking in AI}.
\end{graybox}

\subsubsection{Consistency across devices}

Figure~\ref{fig:pixel_tg_compare} shows that across devices, with the same keyword, text generation results can vary by a significant amount. Qwen had a Spearman $\rho = 0.261$, and SmolLM had $\rho = 0.697$. This suggests that in text generation, inherent randomness in the open ended prompts can lead to large deviations across devices. 

However, Figure~\ref{fig:pixel_sa_compare} shows that sentiment analysis results have much greater correlation across devices, with Qwen having a Spearman $\rho = 0.939$. 

\begin{graybox}
\underline{Insight 4:} \textit{Due to the open ended nature of text generation prompts, their energy consumption can fluctuate across environments. By contrast, sentiment analysis consumption stays relatively consistent}.
\end{graybox}

\subsection{RQ3: How is response quality affected by prompt keywords, and how does it correlate with energy consumption?}
\subsubsection{Text Generation}

Accuracy is measured on a scale from 0--4, with 2 being a passing score (criteria described above). In text generation, the keyword has little impact on response accuracy, as seen in Figure~\ref{fig:tg_accuracy}. 
The Spearman correlation between the rankings by energy consumption, and the rankings by accuracy average to -0.086 for text generation, nearly random. 

\begin{graybox}
\underline{Insight 5:} \textit{In text generation, optimizing prompt wording for energy efficiency has no noticeable impact on response quality}.
\end{graybox}

\subsubsection{Sentiment Analysis}

In sentiment analysis, however, the correlation was -0.632, which is a moderate inverse correlation. This is also apparent in Figure~\ref{fig:sa_accuracy}, where less efficient keywords have lower accuracy. This inverse correlation suggests longer responses can be prone to errors, while shorter responses answer concisely and accurately. 

\section{Future Work}

Our current study focuses on an empirical analysis of how prompt keywords affect the energy consumption of on-device LLM inference.
While our findings provide initial evidence of a measurable correlation between linguistic structure and power usage, several open questions and opportunities remain for future exploration.

\subsection{Development of a Green Prompt Engine}

We plan to develop a \textit{green prompt engine} capable of automatically rewriting user instructions to minimize energy usage while preserving semantic intent and output quality.
This system would operate at three levels:
\begin{enumerate}
    \item \textbf{Linguistic optimization:} reformulating high-energy prompts using alternative verbs or structures known to reduce token length and reasoning depth.
    \item \textbf{Semantic equivalence assurance:} applying sentence embedding or entailment checks to maintain meaning consistency.
    \item \textbf{Energy feedback loop:} incorporating real-time measurement data to iteratively refine rewriting strategies.
\end{enumerate}
Such a framework could be integrated into mobile assistants or developer toolchains, providing transparent sustainability optimization with minimal user intervention.

\subsection{Cross-Device and Cross-Model Generalization}

Our present experiments focus on a single mobile platform and one or two compact LLM variants.
Future work will extend this to a broader set of devices, including smartphones, embedded NPUs, and edge accelerators such as Jetson or Coral TPUs.
Comparing different architectures and hardware configurations will reveal how linguistic-computational correlations generalize across hardware generations and instruction sets.

\subsection{Integration With Multi-Objective Optimization Frameworks}
This integration could leverage reinforcement learning or neural architecture search (NAS) techniques similar to those in PlatformX \cite{tu2025platformx}, extending the concept from model design to prompt formulation.
Such unification would transform prompt engineering from a heuristic process into a systematic optimization problem for sustainable AI deployment.

\section{Conclusion}

Prompt formulation directly shapes the energy footprint of on-device LLM inference. Our empirical results suggest that prompt engineering is not only a semantic control but also a practical, user-level strategy for improving energy efficiency---paving the way for sustainable and adaptive \textit{green prompt engineering}.

\section*{Limitations}

Several limitations of this study should be acknowledged. First, each test was conducted only once without repetition, which reduces the reliability of measurements and leaves results susceptible to random variation. Second, all models tested used the same quantization scheme, limiting the generalizability of our findings across different quantization levels. Third, the prompt dataset was drawn exclusively from two sources: Alpaca-GPT4 and Yelp reviews, which constrains the diversity and breadth of the evaluation. Fourth, the judge LLM used for response quality evaluation appeared to conflate accuracy with other metrics; for instance, responses were occasionally penalized in relevance scores despite being topically appropriate, potentially amplifying accuracy-related variance and undermining measurement reliability. Finally, the substantial disparity in prefill times observed between the Orange Pi 5 Pro and the Google Pixel 9a raises questions about device-level consistency, suggesting that closer monitoring of resource utilization and expanded testing across a wider range of devices would be warranted in future work.

\bibliography{anthology}

% \printbibliography

\clearpage

\appendix
\onecolumn
\section{GEval Metric Definitions}
\begin{lstlisting}
GEval(
    name="Relevance",
    criteria="Evaluate whether the actual output meaningfully attempts to address the questions, tasks, or instructions expressed in the input prompt.",
    evaluation_steps=[
        "Decompose the input prompt into all explicit questions, tasks, or instructions",
        "Identify any clearly implied or dependent sub-requests required to fulfill the prompt",
        "Check whether the actual output attempts to address each identified component",
        "Penalize missing, ignored, or substituted prompt components",
        "Do NOT evaluate factual correctness, depth, or writing quality-only whether an attempt is made to respond to the prompt"
    ],
    model="gpt-4.1-mini",
    evaluation_params=[LLMTestCaseParams.INPUT, LLMTestCaseParams.ACTUAL_OUTPUT]
),
GEval(
    name="Accuracy",
    criteria="Evaluate whether the factual claims in the actual output are correct, verifiable, and consistent with established knowledge or information explicitly provided in the input.",
    evaluation_steps=[
        "Identify all factual claims made in the actual output",
        "Verify each claim against reliable general knowledge or facts stated in the input",
        "Penalize claims that are clearly false or contradict known facts or the input",
        "Penalize incorrect or incomplete answers to explicit factual questions in the input",
        "Do NOT penalize appropriately qualified uncertainty (e.g., 'may', 'depends') when facts are ambiguous or unspecified",
        "Do NOT evaluate writing style, coherence, verbosity, or relevance-only factual correctness"
    ],
    model="gpt-4.1-mini",
    evaluation_params=[LLMTestCaseParams.INPUT, LLMTestCaseParams.ACTUAL_OUTPUT]
),
GEval(
    name="Coherence",
    criteria="Evaluate whether the actual output is logically structured, internally consistent, and easy to follow as a unified response.",
    evaluation_steps=[
        "Identify the main claim, explanation, or narrative the actual output is presenting",
        "Check whether ideas progress in a clear and logical order (e.g., cause-effect, step-by-step, or narrative flow)",
        "Detect internal contradictions or statements that undermine earlier claims and penalize them heavily",
        "Ensure transitions between ideas are clear and do not cause confusion or logical jumps",
        "Penalize tangents or digressions only if they disrupt logical flow or comprehension",
        "Do NOT evaluate factual correctness relative to the input or external knowledge-only internal clarity and consistency"
    ],
    model="gpt-4.1-mini",
    evaluation_params=[LLMTestCaseParams.INPUT, LLMTestCaseParams.ACTUAL_OUTPUT]
),
GEval(
    name="Conciseness",
    criteria="Evaluate whether the actual output communicates the required information efficiently, using no more words than necessary given the intent of the input.",
    evaluation_steps=[
        "Identify the core information or task required by the input prompt",
        "Check whether the actual output includes only information necessary to fulfill that task",
        "Flag redundancy, repetition, filler phrases, or tangential details that do not add meaning",
        "Do NOT penalize length if the input explicitly requests detailed explanation, reasoning, or creativity",
        "Do NOT penalize content that materially improves clarity, correctness, or completeness",
        "Prefer answers that match the shortest clear version an expert would reasonably give"
    ],
    model="gpt-4.1-mini",
    evaluation_params=[LLMTestCaseParams.INPUT, LLMTestCaseParams.ACTUAL_OUTPUT]
)
\end{lstlisting}

\clearpage

\section{Pixel 9a Profiling Lockdown Script}

\begin{lstlisting}
#!/system/bin/sh
# profiling_lockdown.sh
# Pixel 9a profiling lockdown mode (with 80% freq cap)

echo "[*] Entering PROFILING LOCKDOWN MODE"

echo "[CPU] Locking CPU frequencies to ~80% of hardware max"
for cpu in /sys/devices/system/cpu/cpu[0-3]*; do
    cpuinfo_max="$cpu/cpufreq/cpuinfo_max_freq"
    min_file="$cpu/cpufreq/scaling_min_freq"
    max_file="$cpu/cpufreq/scaling_max_freq"
    gov_file="$cpu/cpufreq/scaling_governor"

    if [ -f "$cpuinfo_max" ] && [ -f "$min_file" ] && [ -f "$max_file" ] && [ -f "$gov_file" ]; then
        hw_max=$(cat $cpuinfo_max)
#        target=$((hw_max * 80 / 100))
        target=1328000

        echo userspace > "$gov_file" 2>/dev/null
        echo $target > "$min_file" 2>/dev/null
        echo $target > "$max_file" 2>/dev/null

        echo "$(basename $cpu) locked to $target Hz (~80% of hardware max $hw_max Hz)"
    fi
done
for cpu in /sys/devices/system/cpu/cpu[4-6]*; do
    cpuinfo_max="$cpu/cpufreq/cpuinfo_max_freq"
    min_file="$cpu/cpufreq/scaling_min_freq"
    max_file="$cpu/cpufreq/scaling_max_freq"
    gov_file="$cpu/cpufreq/scaling_governor"

    if [ -f "$cpuinfo_max" ] && [ -f "$min_file" ] && [ -f "$max_file" ] && [ -f "$gov_file" ]; then
        hw_max=$(cat $cpuinfo_max)
#        target=$((hw_max * 80 / 100))
        target=1418000

        echo userspace > "$gov_file" 2>/dev/null
        echo $target > "$min_file" 2>/dev/null
        echo $target > "$max_file" 2>/dev/null

        echo "$(basename $cpu) locked to $target Hz (~80% of hardware max $hw_max Hz)"
    fi
done
for cpu in /sys/devices/system/cpu/cpu[7-7]*; do
    cpuinfo_max="$cpu/cpufreq/cpuinfo_max_freq"
    min_file="$cpu/cpufreq/scaling_min_freq"
    max_file="$cpu/cpufreq/scaling_max_freq"
    gov_file="$cpu/cpufreq/scaling_governor"

    if [ -f "$cpuinfo_max" ] && [ -f "$min_file" ] && [ -f "$max_file" ] && [ -f "$gov_file" ]; then
        hw_max=$(cat $cpuinfo_max)
#        target=$((hw_max * 80 / 100))
        target=2294000

        echo userspace > "$gov_file" 2>/dev/null
        echo $target > "$min_file" 2>/dev/null
        echo $target > "$max_file" 2>/dev/null

        echo "$(basename $cpu) locked to $target Hz (~80% of hardware max $hw_max Hz)"
    fi
done

mali_path="/sys/class/devfreq/20c00000.callisto"
target=560000000
echo "[GPU] Locking GPU frequency to $target Hz"
echo userspace > "$mali_path/governor" 2>/dev/null
echo target > "$mali_path/min_freq" 2>/dev/null
echo target > "$mali_path/max_freq" 2>/dev/null
echo "GPU locked to $target Hz"

# --- Disable Doze / Deep idle ---
dumpsys deviceidle disable
echo "[+] Doze / deep idle disabled"

# --- Thermal throttling: disable all zones ---
for zone in /sys/class/thermal/thermal_zone*; do
    mode="$zone/mode"
    if [ -f "$mode" ]; then
        echo disabled > "$mode" 2>/dev/null
    fi
done
echo "[+] Thermal throttling disabled"

# --- System features ---
pm disable-user --user 0 com.google.android.gms >/dev/null 2>&1
echo "[+] Google Play Services fully disabled"

cmd notification set_dnd none
echo "[+] Push notifications disabled"

settings put system screen_brightness_mode 0
settings put system screen_brightness 0
echo "[+] Adaptive brightness off, screen set to minimum"

settings put secure location_mode 0
echo "[+] Location services disabled"

echo "[*] Lockdown complete. Ready for profiling."

\end{lstlisting}

\clearpage

\section{Sample LLM Prompts}

\begin{lstlisting}
    Text Generation: 

Produce 10 multiple choice questions about the human circulatory system
Produce 2 strategies for reducing stress.
Produce 4 essential questions on the topic of legal regulation of AI.
Produce 5 possible slogans for an online shoe company
Produce a 4 step plan for tackling a problem. Homelessness

Create 10 multiple choice questions about the human circulatory system
Create 2 strategies for reducing stress.
Create 4 essential questions on the topic of legal regulation of AI.
Create 5 possible slogans for an online shoe company
Create a 4 step plan for tackling a problem. Homelessness



    Sentiment Analysis: 

Classify the sentiment of the following Yelp review as positive or negative: Wow!  Yummy, different,  delicious.   Our favorite is the lamb curry and korma.  With 10 different kinds of naan!!!  Don't let the outside deter you (because we almost changed our minds)...go in and try something new!   You'll be glad you did!
Classify the sentiment of the following Yelp review as positive or negative: Amazingly amazing wings and homemade bleu cheese. Had the ribeye: tender, perfectly prepared, delicious. Nice selection of craft beers. Would DEFINITELY recommend checking out this hidden gem.
Classify the sentiment of the following Yelp review as positive or negative: Locals recommended Milktooth, and it's an amazing jewel of Indianapolis. I'm glade I had the chance to experience this.
Classify the sentiment of the following Yelp review as positive or negative: Love going here for happy hour or dinner!  Great patio with fans to beat the StL heat!   Also...very accomodating at this location.  I like the Veal Milanese but with mixed greens instead of pasta!  they'll modify the menu to suit your taste!
Classify the sentiment of the following Yelp review as positive or negative: Great place for breakfast! I had the waffle, which was fluffy and perfect, and home fries which were nice and smashed and crunchy. Friendly waitstaff. Will definitely be back!

Decide whether the sentiment expressed in this Yelp review is positive or negative: Wow!  Yummy, different,  delicious.   Our favorite is the lamb curry and korma.  With 10 different kinds of naan!!!  Don't let the outside deter you (because we almost changed our minds)...go in and try something new!   You'll be glad you did!
Decide whether the sentiment expressed in this Yelp review is positive or negative: Amazingly amazing wings and homemade bleu cheese. Had the ribeye: tender, perfectly prepared, delicious. Nice selection of craft beers. Would DEFINITELY recommend checking out this hidden gem.
Decide whether the sentiment expressed in this Yelp review is positive or negative: Locals recommended Milktooth, and it's an amazing jewel of Indianapolis. I'm glade I had the chance to experience this.
Decide whether the sentiment expressed in this Yelp review is positive or negative: Love going here for happy hour or dinner!  Great patio with fans to beat the StL heat!   Also...very accomodating at this location.  I like the Veal Milanese but with mixed greens instead of pasta!  they'll modify the menu to suit your taste!
Decide whether the sentiment expressed in this Yelp review is positive or negative: Great place for breakfast! I had the waffle, which was fluffy and perfect, and home fries which were nice and smashed and crunchy. Friendly waitstaff. Will definitely be back!



    NOTE: The following prompts produced biased results in the original experiment

Write a 1 sentence summary of digital marketing.
Make 10 sentences with the word 'big'.
Write 7 words related to the word "party".
Create 4 antonyms for the given word Word: Strident
Construct a 3-note melody.


\end{lstlisting}

\end{document}